\documentclass[10pt,twocolumn,letterpaper]{article}

\usepackage[pagenumbers]{cvpr} 

\definecolor{cvprblue}{rgb}{0.21,0.49,0.74}
\usepackage[pagebackref,breaklinks,colorlinks,citecolor=cvprblue]{hyperref}
\usepackage{booktabs}
\usepackage{multirow}
\usepackage{tcolorbox}

\newcommand{\customfootnotetext}[2]{{
\renewcommand{\thefootnote}{#1}
\footnotetext[0]{#2}}}
  
\def\paperID{7889} 
\def\confName{CVPR}
\def\confYear{2024}

\def\OURNAME{RGPT}

\title{RegionGPT: Towards Region Understanding Vision Language Model}

\author{Qiushan Guo\textsuperscript{1},  Shalini De Mello\textsuperscript{2$\dagger$}, Hongxu Yin\textsuperscript{2}, Wonmin Byeon\textsuperscript{2}, Ka Chun Cheung\textsuperscript{2}, \\
Yizhou Yu\textsuperscript{1}, Ping Luo\textsuperscript{1}, Sifei Liu\textsuperscript{2} \\
\textsuperscript{1}The University of Hong Kong
\textsuperscript{2}NVIDIA \\
}

\begin{document}

\twocolumn[{%
\renewcommand\twocolumn[1][]{#1}%
\maketitle
\begin{center}
  \captionsetup{type=figure}
  \vspace{-8mm}
  \includegraphics[width=\linewidth]{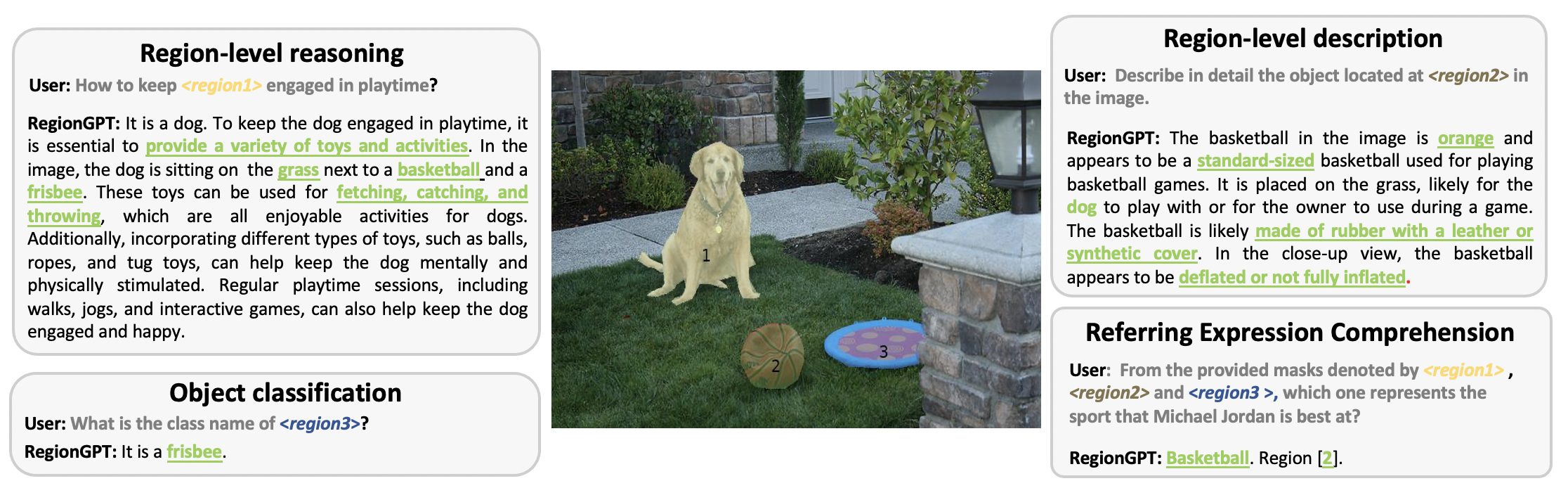}
  \captionof{figure}{
    \footnotesize
    We introduce RegionGPT that enables complex region-level captioning, reasoning, classification, and expression comprehension capabilities for the multimodal large language model. Users can input regions of interest of any shape, utilizing $\langle$region$\rangle$ as a placeholder within the instruction at any position. Such placeholders are subsequently replaced with semantic region-level embeddings that are fed into the language decoder. Best viewed in color.
  }
  \label{fig:case1}
\end{center}%
}]

\customfootnotetext{*}{Qiushan Guo was an intern at NVIDIA during the project. ${\dagger}$ equal contribution. }

\begin{abstract}
Vision language models (VLMs) have experienced rapid advancements through the integration of large language models (LLMs) with image-text pairs, yet they struggle with detailed regional visual understanding due to limited spatial awareness of the vision encoder, and the use of coarse-grained training data that lacks detailed, region-specific captions.
To address this, we introduce RegionGPT (short as RGPT), a novel framework designed for complex region-level captioning and understanding. RGPT enhances the spatial awareness of regional representation with simple yet effective modifications to existing visual encoders in VLMs. We further improve performance on tasks requiring a specific output scope by integrating task-guided instruction prompts during both training and inference phases, while maintaining the model's versatility for general-purpose tasks. Additionally, we develop an automated region caption data generation pipeline, enriching the training set with detailed region-level captions. 
We demonstrate that a universal RGPT model can be effectively applied and significantly enhancing performance across a range of region-level tasks, including but not limited to complex region descriptions, reasoning, object classification, and referring expressions comprehension. Code will be released at the \href{https://guoqiushan.github.io/regiongpt.github.io/}{project page}.
%
\end{abstract}    
\vspace{-2mm}
\section{Introduction}\vspace{-2mm}
\label{sec:intro}
Vision Language Models (VLMs) have marked a notable convergence between visual and linguistic domains in artificial intelligence. With the emergence of Multimodal Large Language Models (MLLMs) \cite{liu2023visual,liu2023improved,zhu2023minigpt,li2023blip,instructblip,alayrac2022flamingo,anil2023palm}, there has been a notable enhancement in the field's ability to interpret images and streamline interactions between humans and VLMs. 
However, despite their effectiveness in understanding entire images, these models still struggle with analyzing specific regions in detail. On the other hand, fine-grained understanding is vital for advanced vision tasks, including the analysis of object attributes and the interpretation of inter-object relations.

Addressing region-level complex understanding in VLMs demands the alignment of spatial information and semantics. 
To acheive this, existing works \cite{peng2023kosmos, chen2023shikra, zhu2023minigpt, liu2023improved} learn inputting regions of interest in textual form, \eg [$x_1$,$y_1$,$x_2$,$y_2$], which share the same model structure as that used for image-level tasks. However, this relies heavily on the language decoder to interpret the position, inadvertently overlooking the prior positional information provided by the visual encoder. Such an oversight can lead to a gap in effectively integrating visual cues with linguistic context, which is crucial for tasks involving detailed image understanding. In a more advanced approach, GPT4RoI \cite{zhang2023gpt4roi} introduces spatial boxes with RoI-aligned features, training the model specifically on region-text pairs. Despite that, the positional format is restricted to a box. And yet the potential for region-specific visual representation, which could offer more expressive fine-grained details and hence benifit downstream vision tasks, remains under-explored.

In this paper, we present \OURNAME, a general framework designed to facilitate complex region-level captioning and understanding. Specifically, we discover that simply refining the visual features extracted by CLIP and employing Mask Pooling to accommodate regions of interest (RoI) of any shape significantly enhances the language model performance on understanding spatial-aware semantic concepts.
Furthermore, we develop task-guided instruction prompts that seamlessly integrate the vision tasks, such as closed-set classification and referring expression comprehension, into our framework. This is achieved by specifying these tasks with visual question answering and response formats.
Existing available region-level captioning datasets, such as ReferCOCOg \cite{kazemzadeh2014referitgame} and VG \cite{krishna2017visual}, tend to provide overly simplistic descriptions of regions, lacking detailed attributes such as color, shape, style and their spatial relation with the surroundings.
To reduce the burden of manual labeling, we propose an automated pipeline for annotating detailed region-level captions, which is achieved by reformatting the existing object detection dataset and employing a two-stage GPT-assisted approach.
Our annotated captions average 87.14 words per region, substantially surpassing the 8.46 words in ReferCOCOg, thereby providing richer contextual information for each region.

Our contributions are threefold: 
(1) We propose \OURNAME, a general framework that harnesses the capabilities of LLMs to tackle complex region-level captioning and understanding tasks. \OURNAME\ is designed for open-ended vision questions, catering to both image-level and region-level tasks.
(2) We design task-guided instruction prompts to specify the output format, thereby eliminating ambiguities in the responses. By transforming vision tasks into VQA tasks, the output patterns are aligned to the language model.
(3) We present a novel data reformation
approach and pipeline, leveraging GPT-assistant, to create high-quality, detailed region-level captions. Our approach significantly enhances the descriptive richness of these captions, with an average word count of 87.14 words per caption.
\begin{figure*}
\begin{center}
    \includegraphics[width=1\textwidth]{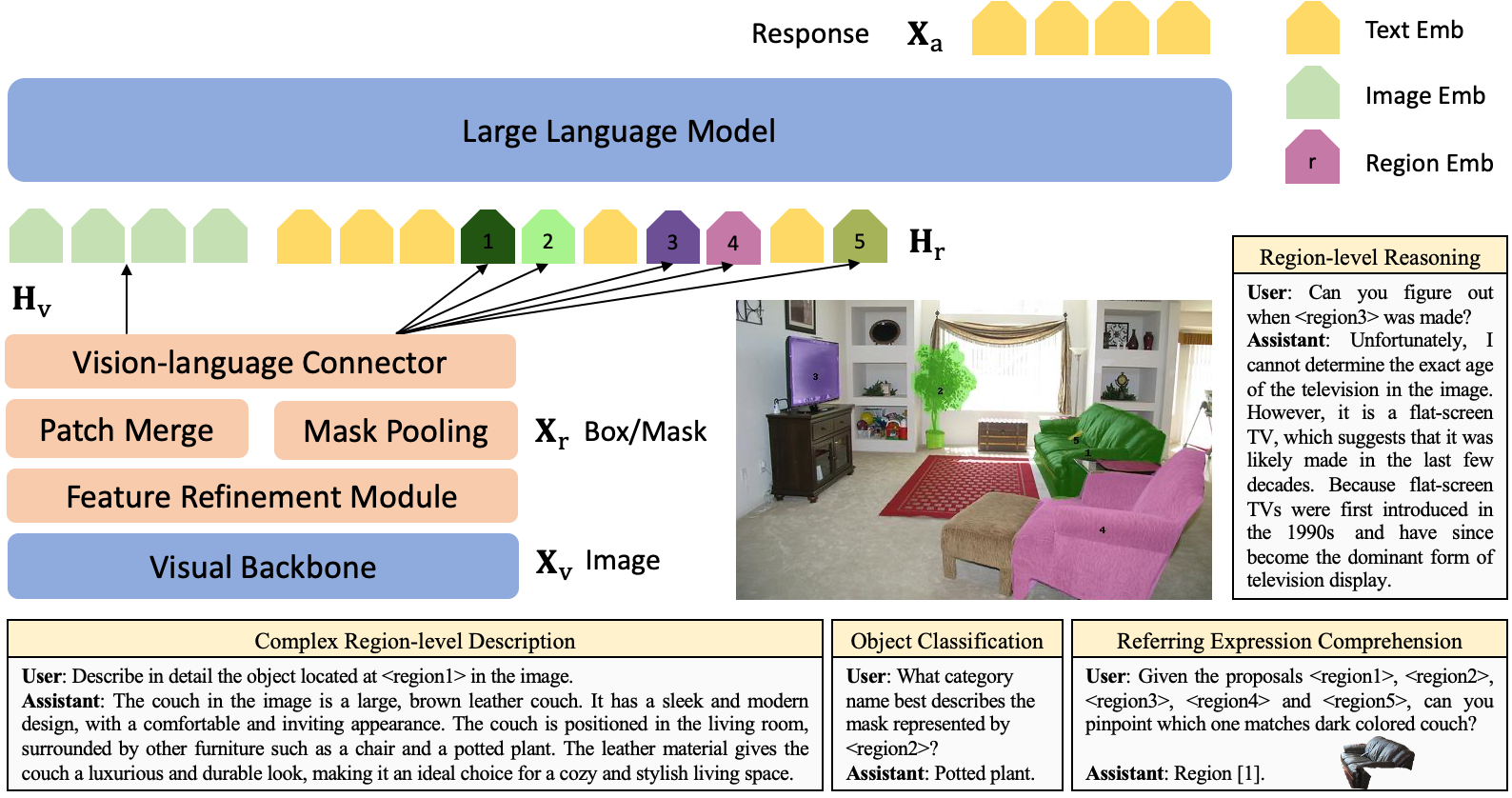}\vspace{-5mm}
\end{center}
\caption{\textbf{Overview of the proposed \OURNAME\ architecture.} Starting from a visual backbone, we extract low-resolution semantic features from an input image $X_v$. Then, a feature refinement module is composed to obtain higher-resolution feature maps. With a patch merge module, the feature maps are further merged to reduce the length of input image-level sequence. The mask features are obtained by averaging the feature in the target region $X_r$, inputted as another branch, with Mask Pooling layer. Both the image-level feature and region-level feature share the connector for semantic consistency. The example interactions demonstrate the model's capabilities in complex region-level description, reasoning, object classification, and referring expression comprehension.}
\vspace{-3mm}
\label{Fig:overview}
\end{figure*}

\section{Related Work}\vspace{-1mm}
\label{sec:Relate}

\subsection{Large Language Model}
Large Language Models have recently gathered considerable interest in the realm of Natural Language Processing (NLP), which is viewed as a form of artificial general intelligence. This surge in attention is attributable to their remarkable proficiency in several key areas: language generation, in-context learning, and the integration of extensive world knowledge and reasoning abilities. The early potential of LLM was first showcased by groundbreaking works such as, BERT \cite{devlin2018bert} and GPT \cite{radford2018improving}. This initiated a trend of scaling up that led to a succession of significant advancements, including T5 \cite{raffel2020exploring}, GPT-3 \cite{brown2020language}, Flan-T5 \cite{chung2022scaling}, PaLM \cite{chowdhery2022palm}, among others. As training data and model parameters expanded, this scaling-up progress culminated in the development of ChatGPT \cite{schulman2022chatgpt} by OpenAI. ChatGPT, leveraging a generative pre-trained model and refined through instruction tuning \cite{ouyang2022training} based on human feedback, demonstrates unparalleled capabilities in engaging in human-like conversations. Rapid advancements in open-source LLMs, such as Llama \cite{touvron2023llama}, Llama-2 \cite{touvron2023llama2} and Vicuna \cite{vicuna2023}, have also started to make them increasingly competitive with ChatGPT.

\subsection{Multimodal Large Language Model}

LLMs have demonstrated formidable capabilities in prior knowledge and reasoning, prompting interest in other modalities. This has led to efforts aimed at extending LLMs into the multimodal domain, where they can interact with and interpret information across various inputs beyond just text. 
For image modality, end-to-end instruction tuning on image-text pairs is proposed to connect the visual backbone with language decoder. Flamingo \cite{alayrac2022flamingo}, BLIP-2 \cite{li2023blip}, LLaVA \cite{liu2023visual} and MiniGPT4 \cite{zhu2023minigpt} are the pioneers to train vision-language connector or language decoder on image-level vision tasks, such as image captioning and visual question answering. Inspired by these pioneers, more recent works are emerged to construct user-friendly interaction dataset \cite{gong2023multimodal, li2023otter} and lightweight trainable weights \cite{zhang2023llama, gao2023llama}. Some other interesting works have made remarkable progress by extending LLM to audio \cite{huang2023language, chen2023x}, medical VQA \cite{zhang2023pmc,moor2023med}  and  control systems \cite{driess2023palm,mu2023embodiedgpt}.

\subsection{Region-level Vision Language Model}
Traditional region-level tasks are common practice in computer vision, such as object detection \cite{ren2015faster, carion2020end}, instance segmentation \cite{he2017mask} and semantic segmentation \cite{ronneberger2015u}, which aims at localizing the regions of interest and close-set classification. Open-vocabulary region-level recognition tasks \cite{xu2022groupvit, xu2023open} target at understanding an object with arbitrary categories described by texts. Recently, region-aware MLLMs, like KOSMOS-2 \cite{peng2023kosmos}, Shikra \cite{chen2023shikra}, MiniGPT-2 \cite{chen2023minigpt} and LLaVA \cite{liu2023improved}, learn inputting regions information in textual form, which heavily rely on the language decoder to interpret position. We argue that incorporating a visual spatial-aware module can extract region-level features more directly and efficiently. By utilizing a visual-language connector, these features enable the complex region-level captioning and reasoning ability. VisionLLM \cite{wang2023visionllm}, GPT4RoI \cite{zhang2023gpt4roi} and ASM \cite{wang2023all} utilize spatial boxes with ROI-aligned features to align the region-level features into LLM word embedding space. However, the input positional format is restricted to a box. Besides, the region visual representation for fine-grained details remains under-explored. On the contrary, our model supports any-shape region as input and focuses on complex reasoning and captioning. Meanwhile, we introduce task-guided instruction prompts to transforming vision tasks into VQA tasks, whose output patterns are aligned with the language model.

\section{Method}

\OURNAME\ is a multimodal large language model with strong capabilities in understanding and referring to specific regions. It can take a inputs of any 2D region, usually in the form of a box or a mask, and provide answers based on instructions. By setting rules for how it should respond to instructions, the model is able to output in a useful and consistent format. This feature allows \OURNAME\ to classify objects at the region level in a closed vocabulary. Additionally, by giving the model region proposals, it can identify specific objects or regions given the query description.
This makes \OURNAME\  a practical tool for tasks that require detailed understanding and processing of different regions within an image.

\begin{table*}[!h]
\centering
\resizebox{0.98\linewidth}{!}{
\begin{tcolorbox}[colback=white!100] {
\centering

{\begin{tabular}{p{1.0\columnwidth} c}
{\bf COCO Object Detection} &\\
\textbf{User}: What category name best describes the region represented by \\
$\langle region1 \rangle$? \textcolor[rgb]{0.8,0,0}{Answer the question using COCO-80 category names.} & 
\hspace{-7.5cm}
\multirow{-3}{*}
{
\includegraphics[height=6.35cm]{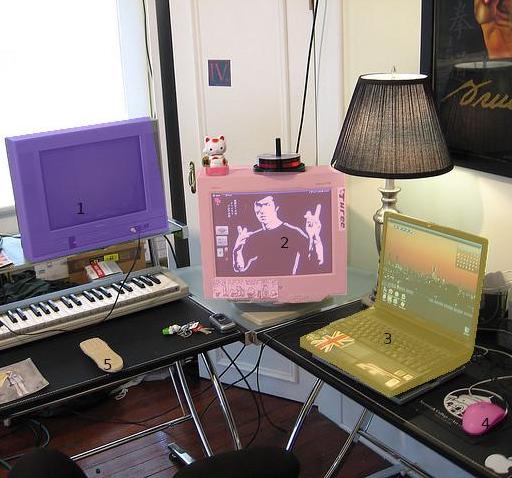}
}
\\
\textbf{Assistant}: \textcolor[rgb]{0.8,0,0.8}{TV} \\
\textbf{User}: How would you label the section $\langle region3 \rangle$? \\
\textbf{Assistant}: \textcolor[rgb]{0.9,0.8,0.5}{Laptop} 

\rule{9.5cm}{1pt} & \\

{\bf Referring Expression Comprehension} &\\
\textbf{User}: From the provided masks denoted by $\langle region1 \rangle$,
$\langle region2 \rangle$, \\
$\langle region3 \rangle$,
$\langle region4 \rangle$ and
$\langle region5 \rangle$, which one fits ``a desktop \\
monitor with Bruce Lee photo on it"? \textcolor[rgb]{0.8,0,0}{Answer the question using} \\
\textcolor[rgb]{0.8,0,0}{the template Region [x].} \\
\textbf{Assistant}: \textcolor[rgb]{0.97265625, 0.55, 0.27265625}{ Region [2] }. \\
\textbf{User}: Match the object with its description: a black Xbox 360. \\
\textbf{Assistant}: \textcolor[rgb]{0.04453125, 0.54921875, 0.14453125}{No corresponding options found.} 
\end{tabular}}}
\end{tcolorbox}
}

\caption{\textbf{Task-guided instruction prompt} to indicate the response format. Two specific tasks are illustrated here. The guided prompt is highlighted \textcolor[rgb]{0.8,0,0}{in red}. We empirically show that instruction prompt is able to adjust the output format and significantly improves the mAP and accuracy on COCO 2017 \textit{val} set.}
\vspace{-5mm}
\label{tab:prompts}
\end{table*}

\subsection{Model Architecture}

An overview of our method \OURNAME, for region-level understanding and image-level understanding is shown in Fig.~\ref{Fig:overview}. It contains an image encoder to extract semantic features, a feature refinement module for the refinement of the low-resolution feature map, an MLP layer to project visual features into the word embedding space and a large language model taking both visual and text tokens.

\noindent\textbf{Visual Backbone.} \OURNAME\ adapts a pretrained CLIP ViT-L \cite{radford2021learning} model as the visual backbone. The visual backbone is frozen during the entire training process. Specifically, an input image $\mathbf{X}_{\mathrm{v}}$ is encoded into a low-resolution feature map $\mathbf{Z}_{\mathrm{LRes}} = f(\mathbf{X}_{\mathrm{v}})$ by the visual backbone.

\noindent\textbf{Feature Refinement Module.} The visual backbone yields a low-resolution feature map, which is not capable of representing small-scale regions and objects. To further refine the visual features, we introduce two deconvolution layers of stride 2 to produce feature maps up-scaled by $4\times$, i.e., 
$\mathbf{Z}_{\mathrm{HRes}} = g(\mathbf{Z}_{\mathrm{LRes}})$.
Our method aims to understand any arbitrary-shaped region of the image, therefore, we choose Mask Pooling to extract region-level features from the high-resolution feature map. More concisely, we average the features of $\mathbf{Z}_{\mathrm{HRes}}$ in region $\mathbf{X}_{\mathrm{r}}$ to get the region-level feature $\mathbf{Z}_{\mathrm{r}} = \mathrm{MaskPool}\ (\mathbf{Z}_{\mathrm{HRes}}, \mathbf{X}_{\mathrm{r}})$.

Since the visual features are flatten as sequence input to language decoder, therefore, high-resolution feature map gets longer sequence input
, which significantly lowers the training and inference efficiency. Hence, we simply use adaptive pooling layer \cite{liu2018path} to merge image feature patches for image-level feature $\mathbf{Z}_{\mathrm{v}} = \mathrm{AdaPool}\ (\mathbf{Z}_{\mathrm{HRes}}, (\text{H}, \text{W}))$, where $(\text{H}, \text{W})$ is the target shape of the low-resolution output feature map.

\noindent\textbf{MLP Vision-language Connector.} To project visual features from the visual backbone into the language model's word embedding space, a two-layer MLP is adopted as the vision-language connector. The embedding of a full image is represented as $\mathbf{H}_{\mathrm{v}} = h(\mathbf{Z}_{\mathrm{v}})$ and the region embedding is $\mathbf{H}_{\mathrm{r}} = h(\mathbf{Z}_{\mathrm{r}})$.
Both the image-level and region-level features share the same connector for semantic consistency.

\noindent\textbf{Large Language Model.} \OURNAME\ incorporates Vicuna (7B) \cite{vicuna2023} as the language decoder. Textual inputs are first tokenized and transformed into word embeddings. Both image-level and region-level features, after being processed through the MLP connector, are directly input into the language decoder.

\begin{figure*}
\begin{center}
    \includegraphics[width=1\textwidth]{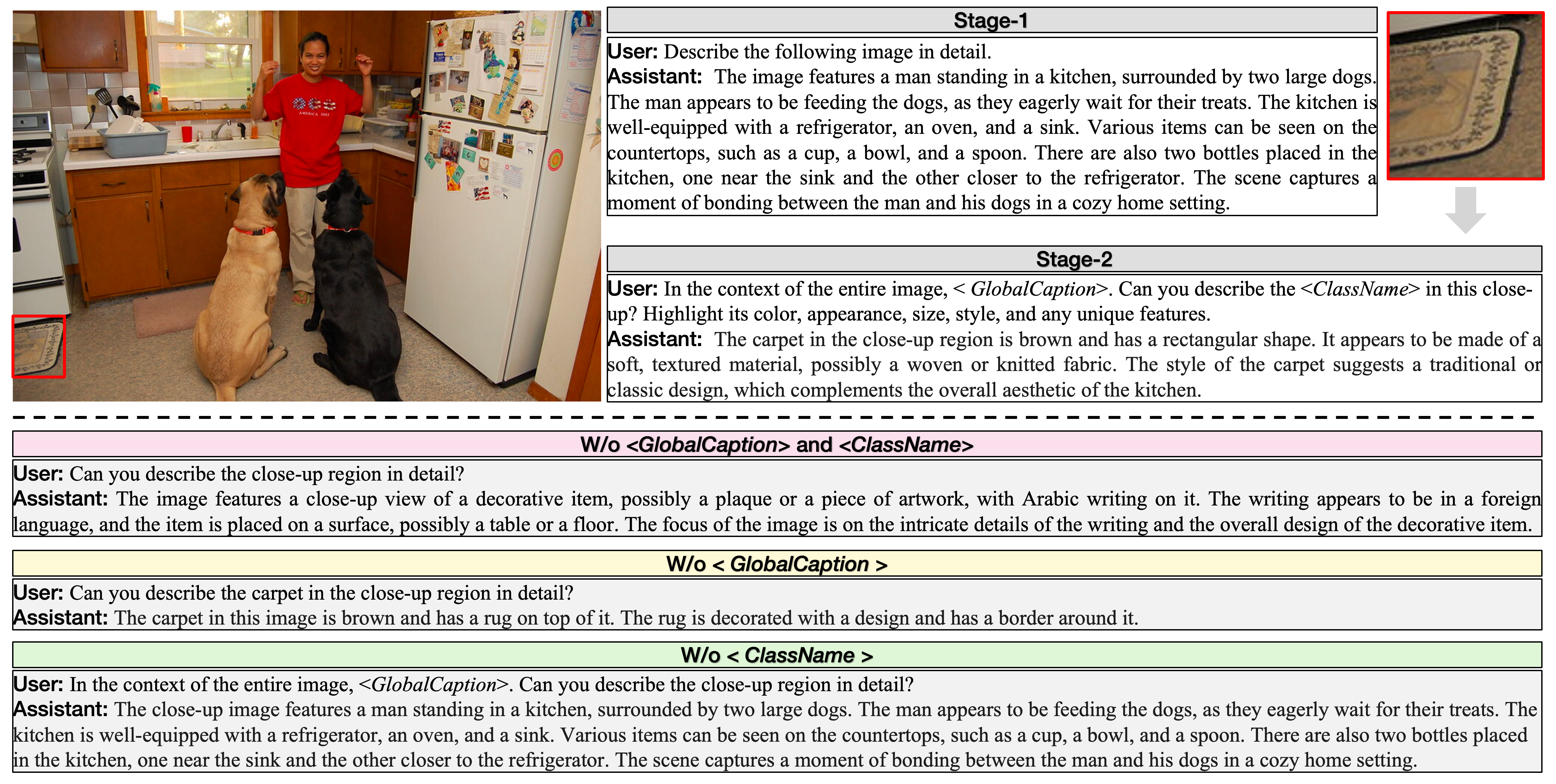}\vspace{-5mm}
\end{center}
\caption{\textbf{Overview of the GPT-assisted region caption generation.} In the upper block, we show our two-stage paradigm in which the final output from the assistant accurately described the local region in terms of color, size and style. In contrast, without the global caption and/or the class name, the assistant either generates vague or over-simplified description, or fails to focus on the region but instead repeating the global context.}
\vspace{-3mm}
\label{Fig:dataset}
\end{figure*}

\subsection{Region-level Instruction Tuning}
\label{sec:RIT}
\noindent \textbf{General prompt format.}
For each image $\mathbf{X}_{\mathrm{v}}$, we generate multi-turn conversation data ([$\mathbf{X}_{\mathrm{v}}$,$\mathbf{X}_{\mathrm{q}}^{1}$], $\mathbf{X}_{\mathrm{a}}^{1}$, ..., $\mathbf{X}_{\mathrm{q}}^{T}$,$\mathbf{X}_{\mathrm{a}}^{T}$), where $T$ is the number of turns, $\mathbf{X}_{\mathrm{q}}^{t}$ is the $t$-th instruction and $\mathbf{X}_{\mathrm{a}}^{t}$ is the corresponding response, following \cite{liu2023visual}.
The image is always used as the starting input of the first instruction to provide the contextual information. To facilitate region-level responses, we introduce the special token $\langle region \rangle$ as a placeholder in the user input prompt, which will be replaced by the corresponding region embedding $\mathbf{H}_{\mathrm{r}}$. The training loss is the standard auto-regressive training objective. We only set the response as the learning target, ignoring the instruction parts.

\noindent\textbf {Task-guided instruction prompt.}
The language model is trained without imposing restrictions on the range of its outputs, in pursuit of achieving flexibility and adaptability. However, certain tasks demand specific output formats. For instance, in the context of the COCO detection task, when provided with a specified bounding box, the model is required to output only the corresponding class name. This response must be selected from a predetermined set of 80 candidate categories.
To tailor the model's responses to specific tasks, we craft custom instruction prompts to guide the model to a desirable output format, as shown in Tab.~\ref{tab:prompts}.

The task-guided instruction prompt ensures that the model remains both versatile and accurate in its task-specific applications. We empirically show that our carefully-designed instruction prompt significantly improves the mAP result on COCO 2017 \textit{val} set.

\noindent{\bf Pre-training stage.}
To maintain and enhance the model's capability in understanding images at both the global and regional levels, we adopt a joint pre-training strategy encompassing both image-level and region-level tasks. For global image understanding, we utilize the LAION-CC-SBU-558K dataset \cite{liu2023improved}, employing image captioning as a pretext task. In parallel, to bolster the model's proficiency in interpreting and interacting with regional aspects of images, we engage it with tasks derived from datasets like Visual Genome \cite{krishna2017visual}, ReferCOCOg \cite{kazemzadeh2014referitgame}, and V3Det \cite{wang2023v3det}. These datasets are transformed into multi-turn conversational formats, which help the model in region-based relationship understanding, captioning, and classification. 

While training, we keep the visual encoder and the language models' weights frozen, and train the feature refinement module and MLP vision-language connector to align the image features with language embeddings. 

\noindent{\bf Fine-tuning stage.} We only keep the visual encoder weights frozen, and continue to update the feature refinement module, MLP connector and language model weights. 
Our objective is to develop a model capable of advanced region-level captioning and reasoning. However, the complexity of existing datasets like ReferCOCOg and Visual Genome for captioning is insufficient for our needs. To address this gap, we additionally incorporate the GPT-assisted region caption dataset (detailed in Sec.~\ref{sec:pseudo}) into our training regime. Furthermore, we craft task-guided instructive prompts on COCO-2017 and ReferCOCOg {\it train} set to develop the model’s ability for closed-set object classification and understanding of referring expressions, as shown in Tab.~\ref{tab:prompts}.

\noindent{\bf Data Processing.}
To enhance training efficiency, we optimize the V3Det dataset by balancing the number of bounding boxes across each category. During the pre-training phase, we limit to 100 boxes per category, and in the fine-tuning phase, this is further reduced to 10 boxes per category. For the closed-set object classification task on the COCO dataset, we retained 20 boxes per category for fine-tuning. 
In the case of Visual Genome, we randomly sampled up to 10 boxes per image to generate dialogues. This filtering process is employed to generate dialogues that are rich in diversity and complexity. 
Although this filtering approach reduces the data's volume, it is important to note that both the visual backbone and the language model have already been pre-trained on large-scale datasets. The strong prior knowledge allows the model to perform effectively even with a smaller, yet diverse set of data. Our data processing strikes a balance between training efficiency and robust model performance.

\subsection{GPT-assisted Region Caption Generation}
\label{sec:pseudo}
In this section, we present a GPT-assisted dense caption generation pipeline, developed to construct the Region Caption Dataset (RecapD). Distinct from traditional image-text pair datasets that typically offer a holistic description of images, RecapD provides in-depth annotations focusing on specific object regions within images. These descriptions emphasize attributes such as color, shape, material, and the spatial relationships between objects. The primary objective of RecapD is to address the challenges associated with region-level understanding and referencing in images, thereby significantly enhancing the capabilities of vision language models in detailed visual comprehension.

\noindent{\bf A two-stage approach.} We explore using an existing global-level image captioning VLM, i.e., LLaVA~\cite{liu2023visual} for region-specific tasks. A naive approach is to crop the region of interest (RoI) and adjust it to fit the model's input format. However, this method often leads to inaccurate captions due to the lack of contextual information from the image's surrounding areas. The absence of surrounding information also makes it infeasible for conveying spatial relationships between objects.

Alternatively, we work around the limitation of the VLMs, which does not support the simultaneous input of both global images and local region patches. To circumvent this, in the first stage, we generate a global-level caption for the image using the VLM. This global description is then used as contextual information, which we include in the form of text at the beginning of the prompt. Subsequently in the second stage, by inputting the ROI, the VLM is prompted to describe the specific region represented by the image patch. We illustrate this approach with a detailed example in the following:
\begin{center}
In the context of the entire image, \textit{$<$GlobalCaption$>$}, describe the close-up region in detail.
\end{center}
Remarkably, our observations reveal that even with this two-stage approach, the model often struggles to accurately describe the input region. This inaccuracy largely stems from its inability to correctly identify the object classes within the cropped region. Therefore, we further enhance our approach by incorporating human-annotated class names as an additional condition when prompting the VLM to describe the properties of the region:
\begin{center}
In the context of the entire image, \textit{$<$GlobalCaption$>$}, describe the \textit{$<$ClassName$>$} in the close-up region in detail.
\end{center}

\begin{table}[t]
\centering
\resizebox{\linewidth}{!}{
\begin{tabular}{lccc}
\toprule
Dataset & Images & Regions & Average words \\
\midrule
ReferCOCO \cite{kazemzadeh2014referitgame} & 20K & 142K & 3.50 \\
ReferCOCO+ \cite{kazemzadeh2014referitgame} & 20K & 142K & 3.53 \\
ReferCOCOg \cite{kazemzadeh2014referitgame} & 25.8K & 95K & 8.46 \\
VG \cite{krishna2017visual} & 82.4K & 3.8M & 5.09 \\
\midrule
Ours & 213K & 1.5M & \textbf{87.14} \\
\bottomrule
\end{tabular}
}
\caption{\textbf{Comparison of our dataset with available region-level caption datasets.} Our dataset stands out with a significantly higher average word count per region caption compared to other datasets. This richness in detail provides a robust foundation for complex region-level understanding.}
\vspace{-10pt}
\label{tab:dataset}
\end{table}


\noindent{\bf GPT-assisted prompt augmentation.}
To enhance the model's adaptability to various styles and combinations of user inputs, we augmented the input prompts using ChatGPT-4 \cite{openai2023gpt4}. For instance, besides ``describe the image in detail'', one may also ask ``provide a detailed description of the given image'', or ``share a thorough analysis of the image'', etc, in the first stage. To ensure a diverse range of responses, we created ten different versions of input prompts for both stages, as elaborated in the supplementary material. During data generation, one of these ten variations is randomly selected for each stage to promote diversity in the model's responses.

\noindent{\bf Region caption dataset analysis.}
Utilizing our automated annotation pipeline, we annotate a corpus of 213K V3Det images \cite{wang2023v3det}, leveraging its comprehensive object bounding boxes and class names. This dataset includes about 13,000 precisely labeled concepts, providing a rich foundation for model training. This extensive and precise labeling enhances the reliability of the generated data. To further refine our dataset, we utilize the CLIP model \cite{radford2021learning} to calculate the similarity between the image regions and the corresponding generated region captions. This process allows us to filter out noisy or irrelevant samples, ensuring that only high-quality data is used for training. As shown in Tab.~\ref{tab:dataset}, our dataset is distinguished by having a notably higher average number of words, 87.14 words per caption, in each region's caption versus other datasets. This detailed richness lays a solid groundwork for an in-depth understanding at the region level.

\section{Experiments}
In this section, we present experimental settings and results. The experiments are primarily conducted on region classification \cite{lin2014microsoft}, captioning \cite{kazemzadeh2014referitgame, krishna2017visual}, expression comprehension \cite{kazemzadeh2014referitgame} and object hallucination benchmark \cite{li2023evaluating}. We present both quantitative and qualitative results.

\begin{table}[t]
\centering
\resizebox{\linewidth}{!}{
\begin{tabular}{l | c c c c c c}
\toprule
Methods & PT & IT & Vision & LLM & mAP & Acc (\%) \\
\midrule
CLIP \cite{radford2021learning} & - & - & ViT-L & - & 58.9 & - \\
RegionCLIP \cite{zhong2022regionclip} & - & - & R50x4 & - & 58.3 & - \\
LLaVA$^\dag$ \cite{liu2023visual} & 595K & 158K & ViT-L & Vicuna-7B & - & 40.04 \\
Shikra$^\dag$ \cite{chen2023shikra} & 600K & 5.5M & ViT-L & Vicuna-7B & - & 53.91 \\
GPT4RoI$^\dag$ \cite{zhang2023gpt4roi} & 266K & 731K & ViT-L & LLaVA-7B  & - & 64.01 \\
PVIT$^\dag$ \cite{chen2023position} & 13.7M & 243K & ViT-L + R50x4 & LLaVA-7B & - & 64.53 \\
ASM \cite{wang2023all} & $\sim$22M & $\sim$22M &  ViT-L & Hasky-7B & 69.3 & - \\
{\bf Ours} & 923K & 953K & ViT-L & Vicuna-7B & {\bf 70.0} & {\bf 80.61} \\

\bottomrule
\end{tabular}
}
\caption{\textbf{Comparison with Region-level based methods on COCO-2017 {\it val} set.} Following RegionCLIP \cite{zhong2022regionclip} and PVIT \cite{chen2023position}, we report the results of object classification with ground-truth box on COCO {\it val} set. $^\dag$ represents the results are imported from \cite{chen2023position}. - means that the results are not reported in the source paper.}
\vspace{-10pt}
\label{tab:coco_classification}
\end{table}

\subsection{Implementation details} 
During the entire training process, the visual backbone weights remain unchanged. We train the model with an image resolution of 336$\times$336 during both the pre-training and fine-tuning stages. An input image is padded to achieve a square format, if it is not square.
In the pre-training stage, we employ a cosine learning rate scheduler. The maximum learning rate is set at 1e-3, with a weight decay of 0 and a warmup ratio of 0.03. The model is trained with a batch size of 256 for one epoch. In the fine-tuning stage, the maximum learning rate is reduced to 2e-5, and the batch size is adjusted to 128. All other hyperparameters remain the same as the pre-training stage.

\begin{table}[t]
\centering
\resizebox{\linewidth}{!}{
\begin{tabular}{lcccc}
\toprule
\textbf{Model} & \multicolumn{2}{c}{\textbf{RefCOCOg}} & \multicolumn{2}{c}{\textbf{Visual Genome}} \\
& METEOR & CIDEr & METEOR & CIDEr \\
\midrule
GRIT \cite{wu2022grit} & 15.2 & 71.6 & 17.1 & 142.0 \\
SLR \cite{yu2017joint} & 15.9 & 66.2 & - & - \\
Kosmos-2 \cite{peng2023kosmos} & 14.1 & 62.3 & - & - \\
Ours & \textbf{16.9} & \textbf{109.9} & \textbf{17.0} & \textbf{145.6} \\
\bottomrule
\end{tabular}
}
\caption{\textbf{Performance on the region-level captioning task on RefCOCOg and Visual Genome.} We report METEOR and CIDEr metrics, following the image-level caption task.}
\label{tab:regioncap}
\end{table}

\subsection{Quantitative Evaluation}
\noindent \textbf{Region Classification.} 
We first evaluate the object classification ability of our model on COCO-2017 dataset. The mAP and classification accuracy metrics are reported to quantify performance. Our focus is on region recognition, rather than object localization. Therefore, following RegionCLIP \cite{zhong2022regionclip}, we use ground-truth boxes as the input for positional information. Alongside this, we attach task-guided instruction prompts to the general instruction prompt and input only one bounding box for one-turn conversation. If the output does not fall within the predefined candidate categories of the COCO dataset, we simply discard this prediction and categorize it as a misclassification.

\begin{table}[t]
\centering
\resizebox{\linewidth}{!}{
\begin{tabular}{c|cccc|c}
\toprule
Method & MDETR\cite{kamath2021mdetr} & Shikra \cite{chen2023shikra} & Kosmos-2 \cite{peng2023kosmos} & MiniGPT-V2 \cite{chen2023minigpt} & Ours\\
\midrule
val & 81.64 & 82.27 &  60.57 &  84.44 & \textbf{86.44} \\
test & 80.89 & 82.19 &  61.65 & 84.66 & \textbf{86.96} \\
\bottomrule
\end{tabular}
}
\caption{\textbf{REC on ReferCOCOg val and test set} \cite{kazemzadeh2014referitgame}. As RGPT focuses on region-level understanding rather than localization, hence, we highlight the strength of our model in interpreting complex expressions within the context of the provided regions from \cite{zong2023detrs}.}\vspace{-4mm}
\label{tab:rec}
\end{table}

We report the results of VLMs and feature-based vision models, as shown in Tab.~\ref{tab:coco_classification}. For our baseline, we crop the RoI from images, resize them to the input size, and then compare their features with those of the 80 classes in the COCO dataset to select the category with the highest similarity. 
Additionally, we consider other feature-based methods like RegionCLIP \cite{zhong2022regionclip} and ASM \cite{wang2023all}. RegionCLIP pretrains CLIP model to align the CC3M \cite{sharma2018conceptual} region-text pairs in the feature space. 
ASM is trained on approximately 22M images and the features are produced by the language decoder. The other VLMs use textual formats as output. On the COCO dataset, our approach achieves a mAP of 70.0 and an accuracy of 80.86\%, demonstrating our method's effectiveness in constraining output formats and its strong capability in region-level object recognition.

\begin{table}[t]
\centering
\resizebox{\linewidth}{!}{
\begin{tabular}{l|cccc}
\toprule
Arch.  & Deconv & BiLinear  & Deconv + BiLinear & None  \\
\midrule
AP     &  \textbf{66.8}  &   60.9    &   62.7      &     57.7  \\
AP$_s$ &  51.1           &   52.8    &  \textbf{53.8}      &    42.7   \\
AP$_m$ &  \textbf{71.5}           &   70.8    &  71.4      &    65.2   \\
AP$_l$ &  \textbf{78.0}  &   57.9    &    60.3      &  65.4     \\
\bottomrule
\end{tabular}
}
\caption{
\textbf{Ablation study on the feature refinement module.} The object classification results on COCO 2017 \textit{val} set are reported. We use ViT-B/16 from \cite{zhai2023sigmoid} as our visual backbone, whose input size is 512$\times$512.
Deconv represents our two deconvolution layers design for feature maps of scale 4. 
BiLinear indicates the use of bilinear upsampling for scale 16. 
Deconv + BiLinear means bilinear upsampling the Deconv output for scale 16. 
None refers to no module is used.
}
\label{tab:ablation}
\end{table}
\begin{table}[t]
\centering
\resizebox{\linewidth}{!}{
\begin{tabular}{l|cccc}
\toprule
Model  & AP  & AP$_s$ & AP$_m$ & AP$_l$  \\
\midrule
OpenAI ViT-L-336 &  70.0  &   55.7    &    75.5      &  81.5     \\
SigLip ViT-B-512   &  66.8  &   51.1    &   71.5      &     78.0  \\
SigLip ViT-L-384    &  69.5   &   56.8    &  74.1      &    80.2   \\
SigLip ViT-SO400M-384 &  \textbf{71.0}   &   \textbf{57.9}    &  \textbf{76.5}      &   \textbf{ 81.6}   \\
\bottomrule
\end{tabular}
}
\caption{
\textbf{Ablation study on visual backbone.} The object classification results on COCO 2017 \textit{val} set are reported. We use SigLip models from \cite{zhai2023sigmoid} pre-trained on WebLI dataset \cite{chen2022pali} and OpenAI CLIP model \cite{radford2021learning}. The results demonstrate that our method can be further improved with more powerful visual network.
}\vspace{-3mm}
\label{tab:ablation_network}
\end{table}

\noindent \textbf{Region Captioning.} 
We evaluate the region-level captioning ability of our model on the ReferCOCOg \cite{kazemzadeh2014referitgame} and Visual Genome \cite{krishna2017visual}, employing the same evaluation metrics as used for image-level captioning: METEOR \cite{banerjee2005meteor} and CIDEr \cite{vedantam2015cider}. As illustrated in Tab.~\ref{tab:regioncap}, our model surpasses the region-aware VLM, Kosmos-2 \cite{peng2023kosmos}. The results highlight our model's proficiency in accurately generating referring expressions for image regions.

\noindent \textbf{Referring Expression Comprehension (REC).}
We evaluate expression comprehension of our model on the ReferCOCOg dataset. Our method focuses on region-level understanding, rather than object localization. Therefore, we utilize bounding box proposals from \cite{zong2023detrs} as candidate box sets. If the Intersection Over Union between the ground truth box and any of the candidate boxes is less than 0.5, we include the ground truth box in our set of candidates. The results in Tab.~\ref{tab:rec} only highlight the specific strength of our model in understanding complex expressions within the context of the provided regions.

\begin{table*}[t]
\begin{center}
\setlength{\tabcolsep}{3.5pt}
\resizebox{0.92\textwidth}{!}{
\begin{tabular}{l|l|ccccccc}
\toprule
Datasets & Metrics & Ours & Shikra \cite{chen2023shikra} & InstructBLIP \cite{instructblip} & MiniGPT4 \cite{zhu2023minigpt} & LLaVA \cite{liu2023visual} & MM-GPT \cite{gong2023multimodal} & mPLUG-Owl \cite{ye2023mplug}  \\
\midrule
\multirow{5}{*}{Random}&Accuracy ($\uparrow$) & \textbf{87.80} & 86.90 & 88.57 & 79.67 & 86.00 & 50.10 & 53.97 \\
& Precision ($\uparrow$) & 97.75 & 94.40 & 84.09 & 78.24 & 87.50 & 50.05 & 52.07 \\
& Recall ($\uparrow$) & 78.13 & 79.26 &  95.13 & 82.20 & 84.00 & 100.00 & 99.60 \\
& F1 Score ($\uparrow$) & 86.85  & 86.19 & 89.27 & 80.17 & 85.71 & 66.71 & 68.39 \\
& Yes & 41.20 & 43.26 & 56.57 & 52.53 & 48.00 & 99.90 & 95.63 \\
\midrule
\multirow{5}{*}{Popular}& Accuracy ($\uparrow$) & \textbf{87.20} & 83.97 & 82.77 & 69.73 & 76.67 & 50.00 & 50.90 \\
& Precision ($\uparrow$) & 95.44 & 87.55 & 76.27 & 65.86 & 72.22 & 50.00 & 50.46 \\
& Recall ($\uparrow$) & 78.13  & 79.20 & 95.13 & 81.93 & 86.67 & 100.00 & 99.40 \\
& F1 Score ($\uparrow$) & 85.92  & 83.16 & 84.66 & 73.02 & 78.79 & 66.67 & 66.94 \\
& Yes & 40.93 & 45.23 & 62.37 & 62.20 & 60.00 & 100.00 & 98.57 \\
\midrule
\multirow{5}{*}{Adversarial}& Accuracy ($\uparrow$) & \textbf{85.67} &  83.10 & 72.10 & 65.17 & 73.33 & 50.00 & 50.67 \\
& Precision ($\uparrow$) & 91.99 & 85.60 & 65.13 & 61.19 & 69.02 & 50.00 & 50.34 \\
& Recall ($\uparrow$) & 78.13 &  79.60 & 95.13 & 82.93 & 84.67 & 100.00 & 99.33 \\
& F1 Score ($\uparrow$) & 84.50  & 82.49 & 77.32 & 76.05 & 66.32 & 66.67 & 66.82 \\
& Yes & 42.47 & 46.50 & 73.03 & 67.77 & 61.33 & 100.00 & 98.67\\
\bottomrule
\end{tabular}
}

\caption{ \textbf{Results on the object hallucination benchmark using the POPE evaluation pipeline \cite{li2023evaluating}.} Except for our model and LLaVA \cite{liu2023visual}, the other results are obtained from \cite{chen2023shikra}.}
\label{tab:pope}
\end{center}
\vspace{-7mm}
\end{table*}

\noindent \textbf{Ablation Study on Feature Refinement Module.} We study the effect of the feature refinement module on the object classification task. Our motivation for this module is to refine the CLIP visual features for better spatial-aware semantics. Tab.~\ref{tab:ablation} shows that two-deconvolution-layer design significantly outperforms the baseline model (the last column), demonstrating the effectiveness of feature refinement. 
An interesting observation is that the methods of 16x upsampling (BiLinear and Deconv + BiLinear) enhance the accuracy of classification for smaller objects, though it shows a decrease in performance for larger objects. Our approach achieves a superior trade-off between these two aspects. We believe that implementing more complex and carefully designed feature optimization mechanisms could potentially lead to further improvements in performance.

\noindent \textbf{Ablation Study on Visual Backbone.} We study the effect of the visual backbone on the object classification task. The results in Tab.~\ref{tab:ablation_network} demonstrate that the performance on region-level understanding can be further improved by replacing current visual backbone with a more powerful one.

\noindent \textbf{Object Hallucination.} We evaluate object hallucinations, generating objects that are inconsistent with the target images in the descriptions, of our method using the POPE evaluation pipeline \cite{li2023evaluating}, with the results detailed in Tab.~\ref{tab:pope}. Our approach significantly outperforms recent popular image-level VLMs. Given that our baseline model is the LLaVA model, we attribute this performance gain to our region-level instruction fine-tuning strategy, which significantly refines the model's ability to interpret images with greater precision.

\subsection{Qualitative Evaluation}
\begin{figure}
\begin{center}
    \includegraphics[width=0.45\textwidth]{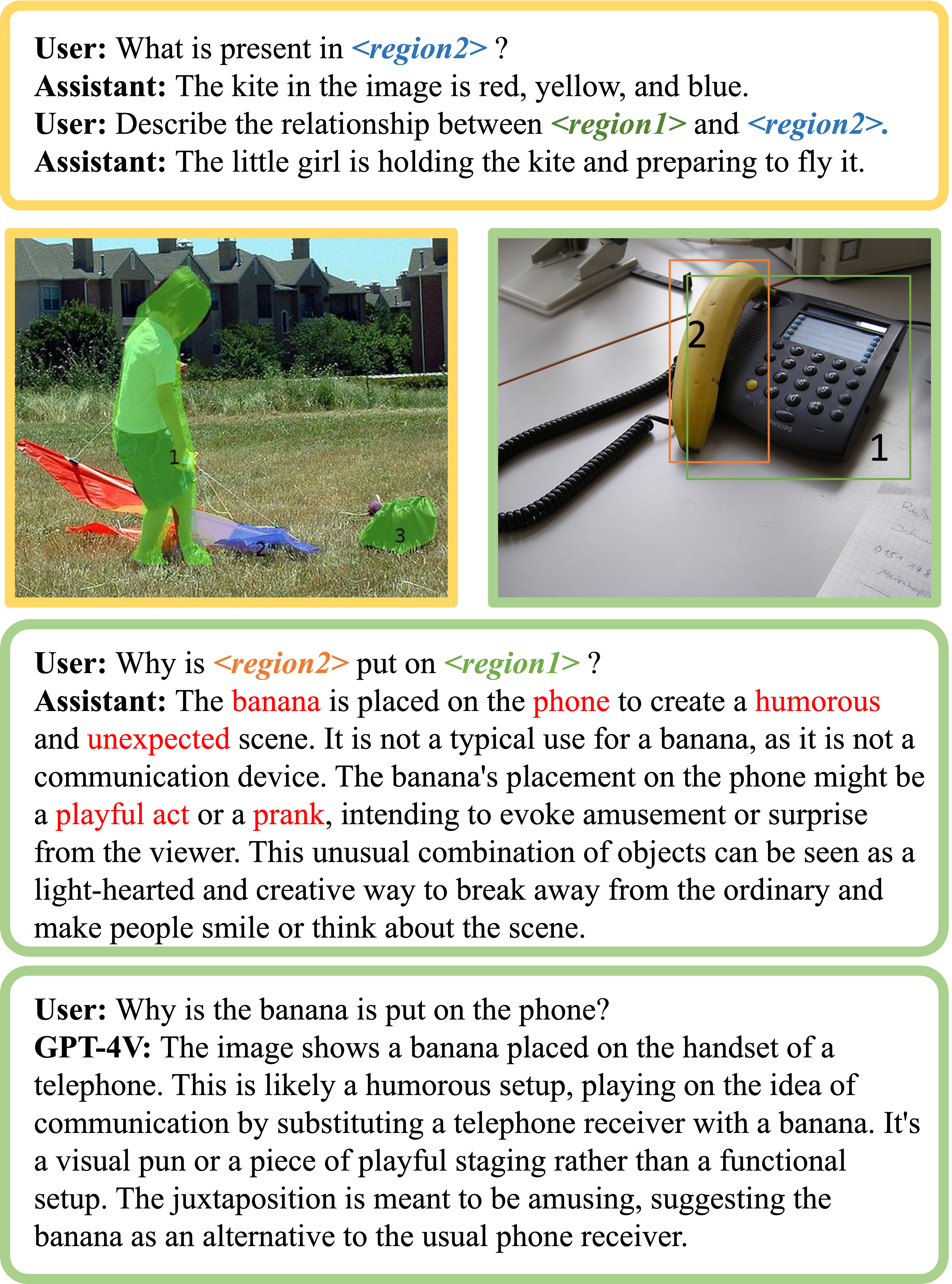}\vspace{-2mm}
\end{center}
\caption{Qualitative evaluation of the mutli-turn conversation of \OURNAME. Our model preserves the mutli-turn conversation and image-level captioning ability.}
\vspace{-15pt}
\label{fig:vis1}
\end{figure}

As demonstrated in Fig.~\ref{fig:vis1}, \OURNAME\ is capable of analyzing of relationships between multiple regions within an image. Additionally, in the green example, our model exhibits complex region-level reasoning abilities akin to those seen in GPT-4V demonstrations, effectively interpreting and explaining visual content.

\section{Conclusion}
In this paper, we present \OURNAME, a general vision-language model that tackles complex region-level captioning and reasoning following user instruction. Our model employs region-level instruction tuning to align the visual feature with the language word embedding space. Besides, we carefully design task-guided instruction prompts to seamlessly blend vision tasks within GPT framework, by converting the vision tasks to VQA tasks and prompting the response format. Finally, we propose a two-stage GPT-assisted annotation pipeline to reformat the object detection dataset and create detailed region-level captions. The results demonstrate that \OURNAME\ achieves impressive performance on the region-level understanding tasks.
{
    \small
    \bibliographystyle{ieeenat_fullname}
    \bibliography{main}
}

\clearpage
\setcounter{page}{1}
\maketitlesupplementary

\renewcommand{\thesection}{\Alph{section}}
\setcounter{section}{0}

\section{Data}
\label{sec:data}
\subsection{Instructions for region-level understanding.} The list of instructions used to briefly describe the region content are shown in Tab.~\ref{tab:prompts_brief_cap}. For detailed region description, the instructions are shown in Tab.~\ref{tab:prompts_detail_cap}. To describe the relationship between the regions, the instructions in Tab.~\ref{tab:prompts_relation_cap} are used. Tab.~\ref{tab:prompts_cls} illustrates the instructions for region classification. For referring expression comprehension, we convert the location task to choice problem, selecting the regions which match the query description.

\subsection{Instruction Tuning Data.}

We list the region-level instruction tuning data in Tab.~\ref{tab:region_pretrain_data} and Tab.~\ref{tab:region_finetune_data} for the Pre-training and Fine-tuning stage. 
For multiple task dataset, we integrate all the instruction-following data into a multi-turn conversation format. This approach enhances training efficiency and ensures the model's capability in multi-round dialogues.

We perform random selection across all annotations for each category, retaining a target number of annotations per category. Images with no annotations selected are discarded.

\begin{table}[ht]
\centering
\resizebox{\linewidth}{!}{
\begin{tabular}{lccc}
\toprule
Pre-train Data              & Size  &   Task &      Random Sampling         \\ 
\midrule
V3Det \cite{wang2023v3det}  & 177K  &   Classification    &  100 per class     \\
VG \cite{krishna2017visual} & 108K   &   Caption \& Relationship      &       No    \\
RefCOCO \cite{kazemzadeh2014referitgame} & 25.8K & Caption \& REC      &       No  \\
\bottomrule
\end{tabular}
}
\caption{Region-level training data in the Pre-training Stage.}
\vspace{-5mm}
\label{tab:region_pretrain_data}
\end{table}

\begin{table}[ht]
\centering
\resizebox{\linewidth}{!}{
\begin{tabular}{lccc}
\toprule
Fine-tuning Data              & Size  &   Task &      Random Sampling         \\ 
\midrule
V3Det \cite{wang2023v3det}  & 98K  &   Classification \& Caption    &  10 per class            \\
COCO \cite{lin2014microsoft}  & 1.5K  &   Classification    &  20 per class            \\
LVIS \cite{gupta2019lvis}  & 52K  &   Classification    &  20 per class            \\
VG \cite{krishna2017visual} & 108K   &   Caption \& Relationship      &       No    \\
RefCOCO \cite{kazemzadeh2014referitgame} & 25.8K & Caption \& REC      &       No  \\
\bottomrule
\end{tabular}
}
\caption{Region-level training data in the Fine-tuning Stage.}
\vspace{-5mm}
\label{tab:region_finetune_data}
\end{table}

\section{More Ablation Studies}
\textbf{Instruction for region classification.}
For the region classification task, we have developed three distinct instruction modes. As shown in Tab.~\ref{tab:coco_eval_prompt}, the first mode involves a one-turn conversation for all RoIs, inputting all RoIs into a single instruction, with the LLM outputting all categories simultaneously. The second mode is a multi-turn conversation for all RoIs, where the LLM conducts multiple rounds of dialogue, classifying one RoI per round. The third mode is a one-turn conversation for one RoI, with the LLM classifying only one RoI per dialogue.

The results in Tab.~\ref{tab:ablation_on_turn} show that multi-turn conversation mode outperforms the other modes, because the previously predicted box provides conditions for the subsequently predicted box, and only one prediction is made at a time, reducing the difficulty.

\begin{table}[ht]
\centering
\resizebox{\linewidth}{!}{
\begin{tabular}{lccc}
\toprule
Mode & One-turn for all RoIs & Multi-turn for all RoIs &  One-turn for one RoI    \\ 
\midrule
mAP & 70.0  &  73.8    &  71.5        \\
\bottomrule
\end{tabular}
}
\caption{\
\textbf{Ablation study on the instruction mode for region classification.} The object classification results on COCO 2017 \textit{val} set are reported. We use ViT-B/16 from \cite{zhai2023sigmoid} as our visual backbone, whose input size is 512$\times$512. All region instances in COCO are used as training data without random sampling.
}
\label{tab:ablation_on_turn}
\end{table}

\noindent\textbf{The number of sample for each concept.}
To assess the impact of the number of annotations per category on classification performance, we conducted experiments on the COCO dataset with varying annotation quantities. We randomly sampled 10, 20, 50, and 200 annotations per category for training. As indicated in the Tab.~\ref{tab:ablation_on_num}, a consistent enhancement in performance was observed with an increasing number of sampled annotations. However, the marginal gain in performance diminished with more data. Notably, increasing annotations from 20 to 200 per category resulted in only a 4 mAP increase.  

\begin{table}[ht]
\centering
\begin{tabular}{lcccc}
\toprule
Num & 10    &   20     & 50     &   200    \\ 
\midrule
mAP & 52.7  &  56.8    &  57.3  &   60.9     \\
\bottomrule
\end{tabular}
\caption{\
\textbf{Ablation study on the annotation quantities for region classification.} The object classification results on COCO 2017 \textit{val} set are reported. We use ViT-B/16 from \cite{zhai2023sigmoid} as our visual backbone, whose input size is 512$\times$512. Different from the other experiments, only the COCO classification region-level data is used to train model.  
}
\label{tab:ablation_on_num}
\vspace{-5mm}
\end{table}

\begin{table}[!h]
\centering
\resizebox{1.0\linewidth}{!}{
\begin{tcolorbox}[colback=white!100] {
\centering

{\begin{tabular}{p{1.0\columnwidth} c}

{
\includegraphics[width=6.85cm]{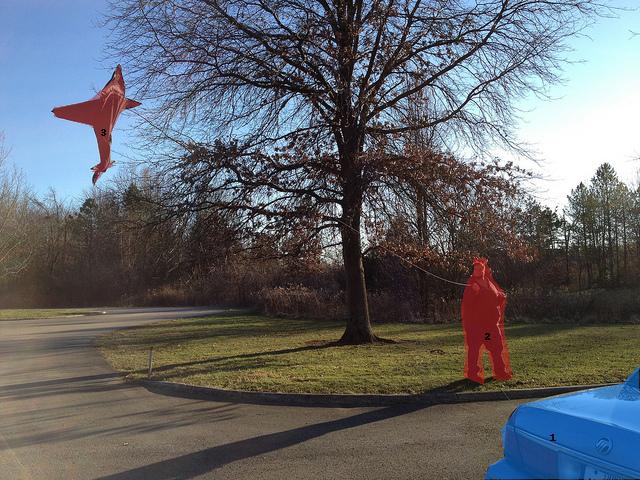}
} \\

{\bf One-turn for all RoIs} &\\
\textbf{User}: What category name best describes the region represented by $\langle region1 \rangle$, $\langle region2 \rangle$ and $\langle region3 \rangle$? \textcolor[rgb]{0.8,0,0}{Answer the question using COCO-80 category names.} 

\textbf{RegionGPT}: Car, person, kite. \\

\rule{7.2cm}{1pt} & \\

{\bf Multi-turn for all RoIs} &\\
\textbf{User}: What category name best describes the region represented by $\langle region1 \rangle$? \textcolor[rgb]{0.8,0,0}{Answer the question using COCO-80 category names.}  \\
\textbf{Assistant}: Car. \\
\textbf{User}: Identify the object or feature present in the region denoted by $\langle region2 \rangle$. \\
\textbf{Assistant}: Person. \\
\textbf{User}: Can you specify the type of object or landscape within the bounds of $\langle region3 \rangle$. \\
\textbf{Assistant}: Kite. \\

\rule{7.2cm}{1pt} & \\

{\bf One-turn for one RoI} &\\
\textbf{User}: What category name best describes the region represented by $\langle region1 \rangle$? \textcolor[rgb]{0.8,0,0}{Answer the question using COCO-80 category names.}  \\
\textbf{Assistant}: Car. \\
\rule{7.2cm}{0.5pt} & \\
\textbf{User}: Identify the object or feature present in the region denoted by $\langle region2 \rangle$. \\
\textbf{Assistant}: Person. \\
\rule{7.2cm}{0.5pt} & \\
\textbf{User}: Can you specify the type of object or landscape within the bounds of $\langle region3 \rangle$. \\
\textbf{Assistant}: Kite. \\

\end{tabular}}}
\end{tcolorbox}
}
\caption{In one-turn for all RoIs mode, all the regions are input to \OURNAME\ within one instruction. For multi-turn for all RoIs, the regions are split and construct sequential conversation. In one-turn for one RoI mode, the conversation is closed for after each instruction.}

\label{tab:coco_eval_prompt}
\end{table}

\begin{table}[!h]
\centering
\resizebox{1.0\linewidth}{!}{
\begin{tcolorbox}[colback=white!100] {
\centering

{\begin{tabular}{p{1.0\columnwidth} c}

{
\includegraphics[width=6.85cm]{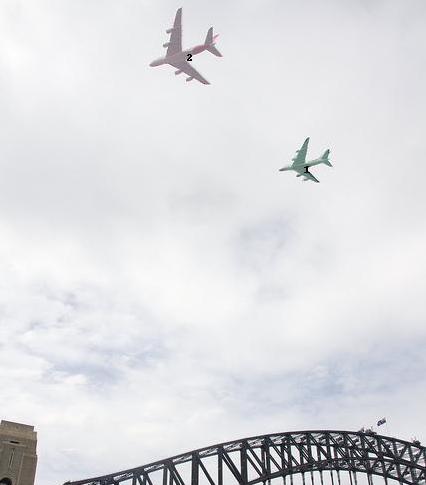}
} \\

{\bf Ground Truth} &\\
\textbf{User}: What category name best describes the region represented by $\langle region1 \rangle$ and $\langle region2 \rangle$? \textcolor[rgb]{0.8,0,0}{Answer the question using COCO-80 category names.} 

\textbf{RegionGPT}: Airplane, airplane. \\

\rule{7.2cm}{1pt} & \\

{\bf Prediction} &\\
\textbf{User}: What category name best describes the region represented by $\langle region1 \rangle$ and $\langle region2 \rangle$? \textcolor[rgb]{0.8,0,0}{Answer the question using COCO-80 category names.} 

\textbf{RegionGPT}: Aerocraft, aerocraft. \\

\end{tabular}}}
\end{tcolorbox}
}
\caption{Task-guided instruction prompt sometimes can not restrict the response format well. The prediction may not be within the COCO-80 classname set.}

\label{tab:limited}
\end{table}

\noindent \textbf{Ablation study on inaccurate annotations.}
Besides mask, RGPT accepts user-drawn bounding box as inputs. While it doesn't directly accept points or scribbles, these can be easily converted into usable masks via SAM. RGPT is robust to inaccurate annotations, like dilated / eroded mask  due to our refinement module, as shown in Tab~\ref{tab:comparison_deconv} and \ref{tab:study_seg_box}.

\begin{table}[h]
\centering
\footnotesize
\begin{tabular}{lcccc}
\toprule
\multirow{2}{*}{Model} & \multicolumn{2}{c}{w/ Feature Refinement} & \multicolumn{2}{c}{w/o Feature Refinement} \\
\cmidrule(lr){2-3} \cmidrule(lr){4-5}
& w/ seg & w/ box & w/ seg & w/ box \\
\midrule
mAP & 71.0 & 70.4 & 65.6 & 60.0 \\
\bottomrule
\end{tabular}
\vspace{-2mm}
\caption{The object classification results on COCO 2017 val set. We use SigLip ViT-SO400M as the visual backbone and input region in the box or mask format.}
\vspace{-2mm}
\label{tab:comparison_deconv}
\end{table}

\begin{table}[h]
\centering
\resizebox{0.9\linewidth}{!}{
\begin{tabular}{lcccc}
\toprule
Model    & Box    &  Seg    &   Seg w\ dilation &  Seg w\ erosion      \\ 
\midrule
mAP      & 69.3   & 70.0    &  71.0             &  68.2  \\
\bottomrule
\end{tabular}
}
\vspace{-2mm}
\caption{The object classification results on COCO 2017 val set. We input box, mask, dilated mask, and eroded mask to RGPT. The kernel is 5 for mask dilation and erosion for 336$\times$336 input.}
\vspace{-6mm}
\label{tab:study_seg_box}
\end{table}
\section{More Qualitative Results}
To demonstrate the multi-turn conversation and complex reasoning capabilities of \OURNAME, we provide more qualitative results. The conversations in Tab.~\ref{tab:supp_stable} show that our model can infer relationship between regions. Tab.~\ref{tab:supp_beach} demonstrates that our model can identify the target from multiple similar objects based on positional cues and perform complex reasoning. Tab.~\ref{tab:supp_motor} illustrates that our model can describe the information of a region in detail, it can even identify the brand of a motorcycle. Tab.~\ref{tab:supp_baseball} indicates that our model can do complex reasoning with multiple regions as input query.

\begin{table*}[!h]
\centering
\resizebox{0.98\linewidth}{!}{
\begin{tcolorbox}[colback=white!100] {
\centering

{\begin{tabular}{p{1.0\columnwidth} c}
\textbf{User}: What can be inferred about $\langle region1 \rangle$ and $\langle region2 \rangle$ relationship? &\\
 
\textbf{RegionGPT}: From the image, it can be inferred that the woman and the & \\
horse share a close relationship, likely as a rider and a horse. The woman & 
\hspace{-6.5cm}
\multirow{-3}{*}
{
\includegraphics[height=4.2cm]{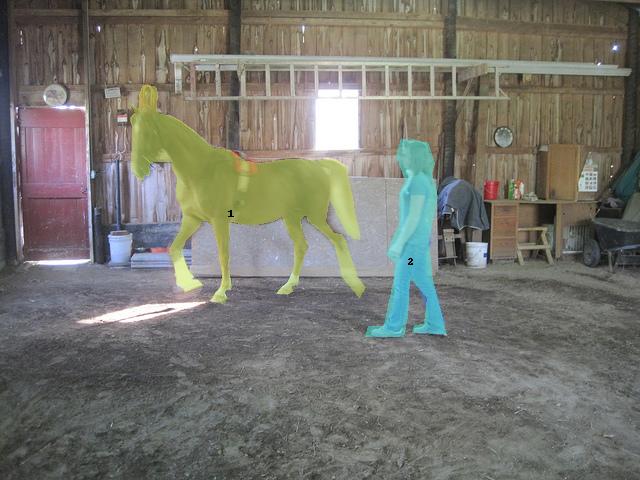}
}
\\
is walking towards the horse, which is standing in a stable, and they are in & \\
a stable or barn setting. This suggests that they might be preparing for a & \\
ride, training session, or simply spending time together. The woman's & \\
presence and interaction with the horse indicate that she is familiar with the & \\
animal and possibly has a bond with it.  & \\
\textbf{User}: What skills might $\langle region2 \rangle$ need for the potential activity? \\
\textbf{RegionGPT}: For the potential activity of horse riding, the woman would \\
need a variety of skills and knowledge to ensure a safe and enjoyable experience. Some of these skills and benefits include: \\
1. Balance and coordination: Horse riding requires maintaining balance and coordination while riding the horse, especially when navigating different terrains or performing various maneuvers. \\
2. Communication and understanding: The woman should be able to communicate effectively with the horse through body language, voice commands, and aids to guide the horse and establish a mutual understanding between them. \\
3. Confidence and assertiveness: Riding a horse can be intimidating for some people, so having confidence and assertiveness in one's abilities is essential to overcome any fears and effectively control the horse. \\
4. Knowledge of horse care and behavior: Understanding the horse's needs, behavior, and body language is crucial for ensuring the horse's well-being and a successful riding experience. \\
5. Safety awareness: Being aware of safety precautions, such as wearing appropriate riding gear, knowing how to mount and dismount the horse, and understanding the horse's temperament, can help prevent accidents and injuries during horse riding. \\
\\
By developing these skills and knowledge, the woman can enjoy a safe and enjoyable horse riding experience while also building a strong bond with her horse.

\end{tabular}}}
\end{tcolorbox}
}

\caption{Qualitative visualization of complex reasoning capability.}

\label{tab:supp_stable}
\end{table*}

\begin{table*}[!h]
\centering
\resizebox{0.98\linewidth}{!}{
\begin{tcolorbox}[colback=white!100] {
\centering

{\begin{tabular}{p{1.0\columnwidth} c}
\textbf{User}: What is $\langle region1 \rangle$ doing, and how can the surroundings affect & \\
his mood? &\\
 
\textbf{RegionGPT}: In the image, a man in a black wetsuit is standing on the & 
\hspace{-7.0cm}
\multirow{-3}{*}
{
\includegraphics[height=4.2cm]{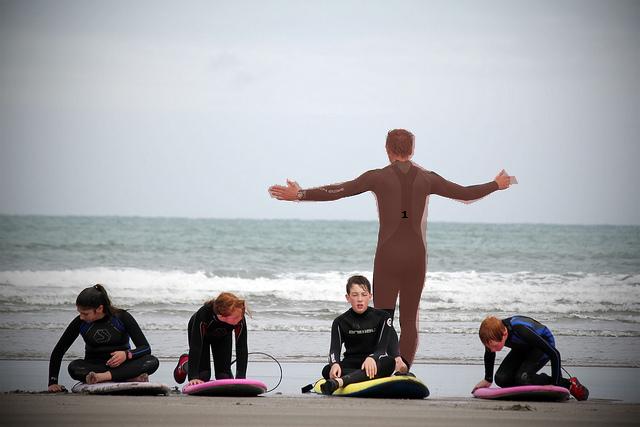}
}
\\
beach, with his arms outstretched. He appears to be enjoying the ocean \\
view and the beach atmosphere. The surroundings, including the \\
presence of other people and surfboards, can positively affect his mood \\
by providing a sense of camaraderie and shared enjoyment of the beach \\
environment. The ocean view and the sound of the waves can also \\
contribute to a relaxing and refreshing atmosphere, which can enhance \\
his overall experience and mood.
\end{tabular}}}
\end{tcolorbox}
}

\caption{Qualitative visualization of complex reasoning capability.}

\label{tab:supp_beach}
\end{table*}

\begin{table*}[!h]
\centering
\resizebox{0.98\linewidth}{!}{
\begin{tcolorbox}[colback=white!100] {
\centering

{\begin{tabular}{p{1.0\columnwidth} c}
\textbf{User}: Describe the $\langle region1 \rangle$ in detail. &\\

\textbf{RegionGPT}: The motorcycle in the image is a black Harley Davidson. \\
It is a large and powerful motorcycle, known for its distinctive design &
\hspace{-7.0cm}
\multirow{-3}{*}
{
\includegraphics[height=4.2cm]{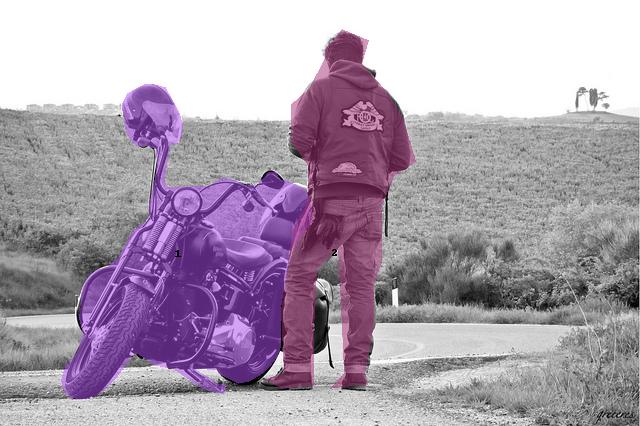}
}
\\
and style. The motorcycle is parked on the side of a road, and it appears \\
to be well-maintained and in good condition. The man standing next to \\
the motorcycle is wearing a jacket, which suggests that the weather \\
might be cool or windy. The motorcycle's size and style make it stand \\
out on the road, and its presence adds a sense of adventure and freedom \\
to the scene. \\
\textbf{User}: What challenge might $\langle region2 \rangle$ face in this environment? \\
\textbf{RegionGPT}: In the image, a man is standing next to a parked motorcycle on a road. The environment appears to be a rural or countryside setting, which might present some challenges for the man. For example, he might face difficulties in finding gas stations, restaurants, or other amenities that are typically found in urban areas. Additionally, the road might be less maintained or less frequently traveled, which could lead to potential hazards or obstacles, such as potholes, debris, or wildlife. The man might also need to be prepared for changing weather conditions, as rural areas can sometimes experience sudden changes in weather or be more prone to natural disasters. Overall, the man would need to be self-sufficient and prepared for the challenges that come with traveling in a rural environment. \\
\end{tabular}}}
\end{tcolorbox}
}

\caption{Qualitative visualization of complex reasoning capability.}

\label{tab:supp_motor}
\end{table*}

\begin{table*}[!h]
\centering
\resizebox{0.98\linewidth}{!}{
\begin{tcolorbox}[colback=white!100] {
\centering

{\begin{tabular}{p{1.0\columnwidth} c}
\textbf{User}: Please tell me the roles of $\langle region1 \rangle$, $\langle region2 \rangle$ and $\langle region3 \rangle$. &\\
Give the reason.  &\\

\textbf{RegionGPT}: The roles of the people in the image are as follows: &
\hspace{-7.0cm}
\multirow{-3}{*}
{
\includegraphics[height=4.2cm]{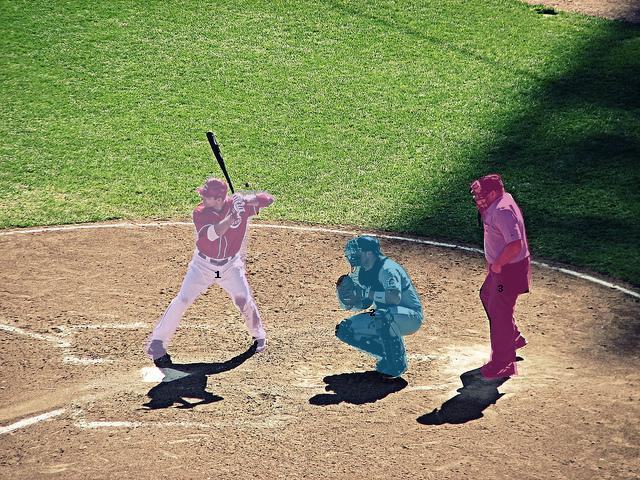}
}
\\
1. Batter: The person holding the baseball bat, who is preparing to hit \\
the ball. \\
2. Catcher: The person wearing a baseball glove, positioned behind the \\
batter to catch the ball. \\
3. Umpire: The person standing behind the catcher, observing the game \\
and making decisions on plays. \\
\\
The reason for these roles is that in a baseball game, the batter's objective is to hit the ball and score runs, while the catcher's role is to catch the ball thrown by the pitcher and prevent the batter from advancing. The umpire's role is to enforce the rules of the game, make decisions on plays, and ensure fair play.

\textbf{User}: What might $\langle region1 \rangle$ aim to achieve during the game? \\
\textbf{RegionGPT}: During the baseball game, the batter, who is holding a baseball bat and standing in the batter's box, aims to achieve a successful hit by hitting the ball thrown by the pitcher. The batter's goal is to hit the ball with enough force and accuracy to reach the bases and ultimately score runs for their team. This can be achieved by using proper batting techniques, timing, and strategy to anticipate the type and speed of the pitch. The batter's performance can significantly impact the outcome of the game, as their ability to hit the ball effectively can contribute to their team's overall success. \\
\textbf{User}: What might $\langle region2 \rangle$ aim to achieve during the game? \\
\textbf{RegionGPT}: The catcher, who is positioned behind the batter in the image, aims to achieve several goals during the game. First, the catcher's primary responsibility is to catch the pitches thrown by the pitcher. They must be prepared to react quickly and accurately to the ball's trajectory to prevent it from getting past them. Second, the catcher plays a crucial role in strategizing with the pitcher, suggesting pitch types and locations to keep the batter off balance and increase the chances of getting them out. Third, the catcher must be ready to throw the ball to the appropriate base to prevent the opposing team's runners from advancing or scoring. Overall, the catcher's goal is to contribute to the team's defensive efforts and help prevent the opposing team from scoring runs. \\
\end{tabular}}}
\end{tcolorbox}
}

\caption{Qualitative visualization of complex reasoning capability.}

\label{tab:supp_baseball}
\end{table*}

\section{Limitations and Future Work}
The current task-guided instruction prompt sometimes can not restrict the response format well. As shown in Tab.~\ref{tab:limited}, for region classification task, the output can be the synonym of ground truth classname. The evaluation of object classification can be reformulated as the semantic similarity between the prediction and ground truth name via a pre-trained text encoder.

\section{Ethics Concerns}
The large language model in our method is pre-trained with the corpus collected by previous works. Despite applying filtering, there may still be potential bias in its response.

\begin{table*}[!h]
\centering
\resizebox{0.98\linewidth}{!}{
\begin{tcolorbox}[colback=white!100] {
\centering
{
\begin{tabular}{p{1.0\columnwidth} c}
    1. Provide a brief caption for the area indicated by $\langle$region$\rangle$. \\
    2. Describe in a short phrase the content within the bounds of $\langle$region$\rangle$. \\
    3. How would you succinctly caption the region highlighted by $\langle$region$\rangle$? \\
    4. Summarize the scene or object present in the section marked by $\langle$region$\rangle$. \\
    5. Can you give a concise description of what's depicted in $\langle$region$\rangle$? \\
    6. Draft a short title for the image content enclosed by $\langle$region$\rangle$. \\
    7. What brief caption would best describe the visual within $\langle$region$\rangle$? \\
    8. Offer a succinct interpretation of the area pointed out by $\langle$region$\rangle$. \\
    9. If you were to provide a short tagline for the content at $\langle$region$\rangle$, what would it be? \\
    10. Give a one-liner description of the region demarcated by $\langle$region$\rangle$. \\
    11. How would you encapsulate the essence of the segment labeled $\langle$region$\rangle$ in a few words? \\
    12. Characterize the content of the image portion specified by $\langle$region$\rangle$ briefly. \\
    13. Craft a mini headline for the visual element spotlighted by $\langle$region$\rangle$. \\
    14. In a few words, how would you narrate the content found within $\langle$region$\rangle$? \\
    15. Pen down a concise caption for the image section delineated by $\langle$region$\rangle$. \\
    16. A short caption of region $\langle$region$\rangle$: \\
    17. A short description of region $\langle$region$\rangle$: \\
    18. A photo containing the region $\langle$region$\rangle$: \\
    19. A region $\langle$region$\rangle$ that shows \\
    20. Write a short description for the region $\langle$region$\rangle$ \\
    21. Write a description for the region $\langle$region$\rangle$ \\
    22. Provide a description of what is presented in the region $\langle$region$\rangle$. \\
    23. Briefly describe the content of the region $\langle$region$\rangle$. \\
    24. Can you briefly explain what you see in the region $\langle$region$\rangle$? \\
    25. Could you use a few words to describe what you perceive in the region $\langle$region$\rangle$? \\
    26. Please provide a short depiction of the region $\langle$region$\rangle$. \\
    27. Using language, provide a short account of the region $\langle$region$\rangle$. \\
    28. Use a few words to illustrate what is happening in the region $\langle$region$\rangle$. \\
    29. Provide an overview of what you see in the region $\langle$region$\rangle$. \\
    30. Can you break down the main elements present in this region $\langle$region$\rangle$? \\
    31. What are the key features or subjects captured in this region $\langle$region$\rangle$? \\
    32. Summarize the primary components of this region $\langle$region$\rangle$. \\
    33. Walk me through the different aspects of this region $\langle$region$\rangle$. \\
    34. Highlight the main points of interest in this region $\langle$region$\rangle$. \\
    35. What stands out to you the most in this region $\langle$region$\rangle$? \\
    36. If you were to give a brief overview of this region $\langle$region$\rangle$, what would you mention? \\
    37. List the primary objects or subjects you identify in this region $\langle$region$\rangle$. \\
    38. Describe the first few things that catch your attention in this region $\langle$region$\rangle$. \\
    39. How would you introduce this region $\langle$region$\rangle$ to someone who hasn't seen it? \\
    40. What are the defining characteristics of this region $\langle$region$\rangle$? \\
    41. Give a concise description of the main content in this region $\langle$region$\rangle$. \\
    42. If you were to caption this region $\langle$region$\rangle$, what might you say? \\
    43. Describe the scene or setting depicted in this region $\langle$region$\rangle$. \\
\end{tabular}
}
}
\end{tcolorbox}
}

\caption{The list of instructions for brief region description.}
\label{tab:prompts_brief_cap}
\end{table*}

\begin{table*}[!h]
\centering
\resizebox{0.98\linewidth}{!}{
\begin{tcolorbox}[colback=white!100] {
\centering
{
\begin{tabular}{p{1.0\columnwidth} c}
    1. Describe in detail the object located at $\langle$region$\rangle$ in the image, including its appearance, style, and any visible details. \\
    2. Provide a comprehensive description of the area marked by $\langle$region$\rangle$, focusing on textures, colors, and any notable features. \\
    3. Elaborate on the artwork shown in the region indicated by $\langle$region$\rangle$, mentioning its color, appearance, size, style, and any standout features. \\
    4. Give a detailed analysis of the scene within the boundary of $\langle$region$\rangle$, touching upon its components, ambiance, and any thematic expressions. \\
    5. Craft a thorough narrative about the piece of the image highlighted by $\langle$region$\rangle$, from its aesthetic qualities to its possible historical context.\\
    6. Explain in depth the characteristics and attributes of the subject found in the segment tagged with $\langle$region$\rangle$. \\
    7. Generate a long, detailed caption for the segment of the image at $\langle$region$\rangle$, covering aspects such as its origin, material, and any symbolic meaning. \\
    8. Paint a vivid picture with words about the region at $\langle$region$\rangle$, diving into the intricacies and nuances present in the area. \\
    9. Zoom in on the area indicated by $\langle$region$\rangle$ and describe every discernible detail, from texture and color to form and function. \\
    10. Offer an expanded description of the contents within the area marked by $\langle$region$\rangle$, encompassing its color, appearance, size, style, and any remarkable features. \\
\end{tabular}
}
}
\end{tcolorbox}
}

\caption{The list of instructions for detailed region description.}
\label{tab:prompts_detail_cap}
\end{table*}

\begin{table*}[!h]
\centering
\resizebox{0.98\linewidth}{!}{
\begin{tcolorbox}[colback=white!100] {
\centering
{
\begin{tabular}{p{1.0\columnwidth} c}
    1. Explain the relationship between the area indicated by $\langle$region$\rangle$ and the region marked by $\langle$region$\rangle$ in terms of their visual or thematic connection. \\
    2. Describe any functional or aesthetic connection between the elements at $\langle$region$\rangle$ and $\langle$region$\rangle$ in the image. \\
    3. Analyze how the region $\langle$region$\rangle$ complements or contrasts with the area $\langle$region$\rangle$ in terms of design and composition. \\
    4. Discuss the interplay between the features located at $\langle$region$\rangle$ and the attributes of the region at $\langle$region$\rangle$. \\
    5. Detail the way in which the area labeled $\langle$region$\rangle$ interacts with or relates to the region designated by $\langle$region$\rangle$ within the image's context.\\
    6. Assess the correlation or disparity between the segment at $\langle$region$\rangle$ and the segment at $\langle$region$\rangle$, including any observable influences or contrasts. \\
    7. Compare the region $\langle$region$\rangle$ with the area $\langle$region$\rangle$ to determine how they either work together or differ substantially within the image. \\
    8. Identify and elaborate on any thematic or stylistic relationships between the contents of $\langle$region$\rangle$ and $\langle$region$\rangle$. \\
    9. Interpret the connection between the area at $\langle$region$\rangle$ and the region at $\langle$region$\rangle$, considering their positions, roles, or symbolism in the image. \\
    10. Clarify how the part of the image within $\langle$region$\rangle$ corresponds with, or is disparate from, the part within $\langle$region$\rangle$ in terms of their visual narrative. \\
\end{tabular}
}
}
\end{tcolorbox}
}

\caption{The list of instructions for region relationship description.}
\label{tab:prompts_relation_cap}
\end{table*}

\begin{table*}[!h]
\centering
\resizebox{0.98\linewidth}{!}{
\begin{tcolorbox}[colback=white!100] {
\centering
{
\begin{tabular}{p{1.0\columnwidth} c}
    1. Identify the object or feature present in the region denoted by $\langle$region$\rangle$. \\
    2. What category best describes the area represented by $\langle$region$\rangle$? \\
    3. Describe the content of the image section highlighted by $\langle$region$\rangle$. \\
    4. Can you specify the type of object or landscape within the bounds of $\langle$region$\rangle$? \\
    5. Which of the following categories best fits the region marked by $\langle$region$\rangle$? Provide your answer.\\
    6. What can you discern from the area indicated by $\langle$region$\rangle$ in the image? \\
    7. Categorize the visual element within the area designated by $\langle$region$\rangle$. \\
    8. Give a brief description of the item or scene captured in the segment marked by $\langle$region$\rangle$. \\
    9. Which classification would you assign to the visual content found at $\langle$region$\rangle$? \\
    10. Determine and describe the primary subject located within $\langle$region$\rangle$. \\
    11. How would you label the section of the image encompassed by $\langle$region$\rangle$? \\
    12. Assess and classify the feature present within the confines of $\langle$region$\rangle$. \\
    13. If you were to tag the section indicated by $\langle$region$\rangle$, what tag would you use? \\
    14. What stands out to you in the region demarcated by $\langle$region$\rangle$? Please classify it. \\
    15. Evaluate the content of the image portion pinpointed by $\langle$region$\rangle$ and provide its category. \\
\end{tabular}
}
}
\end{tcolorbox}
}

\caption{The list of instructions for region category description.}
\label{tab:prompts_cls}
\end{table*}

\begin{table*}[!h]
\centering
\resizebox{0.98\linewidth}{!}{
\begin{tcolorbox}[colback=white!100] {
\centering
{
\begin{tabular}{p{1.0\columnwidth} c}
    1. Given the mask proposals $\langle$region$\rangle$ in the image, can you pinpoint the one that matches $\langle$description$\rangle$. \\
    2. From the provided masks denoted by $\langle$region$\rangle$ in the picture, which one best fits the description of $\langle$description$\rangle$? \\
    3. Looking at the mask suggestions $\langle$region$\rangle$ in the image, identify the one that corresponds to $\langle$description$\rangle$. \\
    4. In the image with mask proposals $\langle$region$\rangle$, please highlight the one that represents $\langle$description$\rangle$. \\
    5. Considering the mask candidates $\langle$region$\rangle$ from the photo, which one would you associate with $\langle$description$\rangle$? \\
    6. Among the mask proposals $\langle$region$\rangle$ in the visual, can you discern the one depicting $\langle$description$\rangle$? \\
    7. From the set of masks labeled as $\langle$region$\rangle$ in the image, which one aligns with the description $\langle$description$\rangle$? \\
    8. Based on the mask data provided as $\langle$region$\rangle$ in the photo, can you spot the one indicative of $\langle$description$\rangle$? \\
    9. In the presented image with mask suggestions $\langle$region$\rangle$, determine which mask resonates with $\langle$description$\rangle$. \\
    10. Given the mask assortment $\langle$region$\rangle$ in the image, please detect the one that matches the characteristics of $\langle$description$\rangle$. \\
    11. Reviewing the mask candidates $\langle$region$\rangle$ from the picture, can you single out the one that fits $\langle$description$\rangle$? \\
    12. From the list of mask proposals $\langle$region$\rangle$ in the image, identify the one that best encapsulates $\langle$description$\rangle$. \\
    13. Considering the provided mask data $\langle$region$\rangle$ in the visual, which one would you say corresponds to $\langle$description$\rangle$? \\
    14. In the snapshot with the mask proposals $\langle$region$\rangle$, please locate the mask that can be described as $\langle$description$\rangle$. \\
    15. Based on the available mask candidates $\langle$region$\rangle$ in the image, can you pick the one that portrays $\langle$description$\rangle$? \\
\end{tabular}
}
}
\end{tcolorbox}
}

\caption{The list of instructions for referring expression comprehension.}
\label{tab:prompts_rec}
\end{table*}

\end{document}